\documentclass[11pt]{article}

\usepackage[preprint]{acl}

\usepackage{times}
\usepackage{latexsym}

\usepackage[T1]{fontenc}

\usepackage[utf8]{inputenc}

\usepackage{microtype}

\usepackage{inconsolata}
\usepackage{longtable}
\usepackage{changepage}
\usepackage{float}
\usepackage{makecell}
\usepackage{array}
\usepackage{enumitem}
\usepackage{graphicx}
\usepackage{dsfont}
\usepackage{fontawesome}
\usepackage{titletoc}
\usepackage{colortbl}
\usepackage{multirow}
\usepackage[most]{tcolorbox}
\usepackage{amssymb}
\usepackage{booktabs}
\usepackage{xcolor}
\usepackage[table]{xcolor}
\usepackage[normalem]{ulem}
\usepackage{array}
\usepackage{hhline}
\usepackage{soul}
\usepackage{lipsum}

\soulregister\citep7
\soulregister\citet7
\soulregister\ref7

\definecolor{gold}{HTML}{FFF3C4}
\definecolor{silver}{HTML}{ECEFF1}
\definecolor{bronze}{HTML}{f0e2d8}

\newcommand{\checkbox}[1]{%
  \ifnum#1=1
    \makebox[0pt][l]{\raisebox{0.15ex}{\hspace{0.1em}$\checkmark$}}%
  \fi
  $\square$%
}

\title{Parameter-Efficient Retrievers for Polish and European Languages}

\author{
\textbf{Sławomir Dadas\thanks{Equal contribution.}},
\textbf{Rafał Poświata\footnotemark[1]},
\textbf{Małgorzata Grębowiec},
\textbf{Michał Perełkiewicz} \\
National Information Processing Institute \\
al. Niepodległości 188b, 00-608 Warsaw, Poland \\
{\faEnvelopeO} \! \texttt{\{sdadas,rposwiata\}@opi.org.pl} \\}

\begin{document}
\maketitle
\begin{abstract}
Dense retrieval systems increasingly rely on multi-billion-parameter language models, whose memory and computational requirements make large-scale indexing, frequent corpus updates, and low-latency serving costly. We present a three-stage training pipeline for developing compact and efficient retrievers that remain competitive with substantially larger models. The pipeline combines cross-lingual alignment, relational knowledge distillation, and contrastive fine-tuning. It requires no original ground-truth relevance labels, relying exclusively on supervision generated by strong embedding models and rerankers utilised as teachers. Using this pipeline, we develop PolDense and EuroDense, both supporting contexts of up to 8,192 tokens. PolDense is a family of six Polish retrievers ranging from 17M to 1B parameters. EuroDense is a 435M-parameter retriever supporting nine European languages. We conduct an extensive evaluation covering 41 Polish and 150 multilingual retrieval tasks. The results demonstrate strong quality-efficiency trade-offs. PolDense-1B outperforms the evaluated retrievers with up to 9B parameters, while the PolDense family forms the Pareto frontier across model sizes. Among the evaluated models below 1B parameters, EuroDense ranks first in both task-averaged and language-averaged performance and leads in seven of nine languages. We release all models publicly.
\end{abstract}

\section{Introduction}

Text retrieval aims to identify passages or documents relevant to a user query within a large corpus. With the widespread adoption of large language models (LLMs), retrieval-augmented generation (RAG) has become a common architecture for practical applications, combining an LLM with a retrieval pipeline that typically includes a retriever and a reranker \citep{lewis2020retrieval}. This demand has driven rapid progress in retrieval research, reflected in benchmarks such as BEIR \citep{thakur2021beir}, MTEB \citep{muennighoff2023mteb}, RTEB \citep{aarsen2025rteb}, and numerous language-specific MTEB suites. Although dense retrieval quality has improved substantially, many top-performing models now rely on multi-billion-parameter LLM backbones. Their computational and memory requirements reduce encoding throughput and increase the cost of corpus indexing, frequent index updates, and low-latency serving, limiting their practicality for collections containing millions of documents.

We introduce \textbf{PolDense} and \textbf{EuroDense}, retrieval models designed to provide a favorable trade-off between quality and deployment cost. PolDense comprises six Polish dense retrievers ranging from 17M to 1B parameters. EuroDense is a 435M-parameter model supporting nine European languages: English, German, French, Spanish, Italian, Portuguese, Dutch, Russian, and Polish. All models support input sequences of up to 8,192 tokens. We train them using a three-stage pipeline consisting of two knowledge-distillation steps followed by contrastive fine-tuning. No stage uses the original ground-truth relevance labels. Instead, training supervision is generated by strong embedding models and a reranker. PolDense establishes a new state of the art while offering high parameter efficiency and outperforming substantially larger retrievers. EuroDense achieves the best aggregate performance among the evaluated multilingual models below 1B parameters. We make our models publicly available\footnote{\url{https://hf.co/collections/OPI-PIB/poldense-and-eurodense}}.

\section{Related Work}
The development of dense text representation techniques has evolved from methods such as Sentence-BERT \citep{reimers-gurevych-2019-sentence} and SimCSE \citep{gao-etal-2021-simcse} to advanced encoder families, including E5 \citep{wang2024textembeddingsweaklysupervisedcontrastive} and BGE \citep{xiao2024bge}. In subsequent years, research focused on developing multilingual embedding models, which include model families such as Multilingual-E5 \citep{wang2024multilingual}, Snowflake-Arctic-Embed \citep{yu2024arctic}, and newer ones like Jina-Embeddings \citep{sturua2025jina, akram2026jina}. Their main advantage is the ability to map multiple languages into a unified vector space, which, for smaller models, enables fast, cost-effective inference in multilingual production environments. However, striving to support dozens of languages within a fixed parameter budget can lead to the phenomenon known as the curse of multilinguality \citep{conneau2020unsupervised, pfeiffer-etal-2022-lifting}, which can result in worse performance on individual languages compared to specialized monolingual models of similar size.
Recent years have seen a trend based on utilising LLMs to create text embeddings, examples of which include E5-Mistral \citep{wang-etal-2024-improving-text}, NV-Embed \citep{lee2025nvembed}, Qwen3-Embeddings series \citep{zhang2025qwen3}, BGE-Multilingual-Gemma2 \citep{chen-etal-2024-m3, xiao2024bge}, or INF-Retriever-v1 \citep{infly-ai_2025}. These models have dominated benchmarks such as MTEB by offering high semantic quality. However, their primary drawback, stemming from parameter counts in the billions, is a significant limitation on their practical application in real-time retrieval systems due to high latency and infrastructure costs.
In the context of the Polish language, existing solutions have predominantly focused on encoder architectures tailored for local retrieval and semantic similarity tasks, including Polish SBERT \citep{dadas2022}, Silver-Retriever-Base-v1 \citep{rybak2024silver}, the MMLW model family  \citep{dadas2024pirb}, and more recent models such as Stella-PL-Retrieval \citep{dadas2024pirb}.

Addressing the need for language-specific efficiency, we introduce dedicated Polish models that offer competitive semantic quality and low latency compared to significantly larger multilingual baselines. Additionally, we propose a multilingual model optimized for nine languages that reduces the risk of parameter degradation and outperforms existing solutions in its size class.

\section{Methodology}
We propose a three-stage training pipeline comprising two knowledge distillation stages followed by retrieval-oriented fine-tuning, which was used to train all of our models. The first two stages use large parallel text corpora for training, while the final stage relies on retrieval datasets consisting of collections of questions and documents. We first describe the objectives employed at each stage, along with our modifications to the methods proposed in the original studies. We then present the model-specific training configurations, including the base and teacher models used, training data, and key hyperparameters.

\subsection{Cross-lingual alignment}

The objective of the first stage is to extend a monolingual embedding model to one or more additional languages while preserving the properties of its original embedding space. This stage builds on the approach proposed by \citet{reimers-gurevych-2020}, in which a teacher model defines a shared semantic embedding space. For each set of parallel sentences, the teacher encodes the source-language sentence, while the student is trained to produce similar embeddings for both the source sentence and its translations. The original method uses mean squared error (MSE) to align the student representations with the teacher representation. The method can also be applied when the teacher and student produce embeddings of different dimensionalities by introducing an additional learned projection layer that maps the student embeddings to the dimensionality of the teacher space. Subsequent studies showed that supplementing or replacing MSE with cosine loss is beneficial for this type of distillation \citep{heffernan-etal-2022-bitext,bocharova-malakhov-2024-ukrainian-embeddings}. We therefore applied a weighted objective combining MSE with cosine distance. The former aligns individual embedding dimensions, whereas the latter encourages agreement between vector directions. For a pair of embeddings $\mathbf{x}_1$ and $\mathbf{x}_2$ the alignment loss is defined as:
\begin{equation}
    \mathcal{L}
    =
    \mathrm{MSE}({x}_1,{x}_2)
    +
    (1-\cos({x}_1,{x}_2))
\end{equation}

\subsection{Relational distillation}

The second stage uses a more advanced distillation objective inspired by \citet{zhang2024jasper}. Its purpose is to transfer not only individual teacher embeddings but also the similarity structure relevant to retrieval. The objective combines three complementary components: \(L_1\), a cosine alignment loss that directly aligns student and teacher embeddings; \(L_2\), a similarity loss that minimizes differences between their within-batch similarity matrices; and \(L_3\), a margin-based relative similarity loss that encourages the student to reproduce the ordering of pairwise similarities determined by the teacher. Together, these objectives preserve the teacher's local embedding geometry and ranking behavior, which are more important for retrieval than matching individual vectors alone. We introduced three modifications to the original method:

\begin{itemize}[wide,labelwidth=0pt,labelindent=0pt,itemsep=0pt,topsep=5pt,parsep=0pt]
\item \textbf{Single-run gradual unfreezing -}
The original method separates training into three stages with different trainable parameter groups and hyperparameters. To simplify the process, we replace them with a single training run governed by a learning-rate scheduler with three consecutive warm-up phases, each lasting 50,000 steps. The first phase activates and warms up the projection layer, the second additionally activates the last four Transformer blocks, and the third activates the remaining model parameters. This preserves gradual unfreezing while using a single training configuration.

\item \textbf{Discardable projection layer -}
In the original method, the projection layer mapping student embeddings to the teacher dimensionality is retained after training, potentially producing very high-dimensional embeddings. Lower-dimensional outputs are obtained through additional projection layers trained using Matryoshka Representation Learning \citep{kusupati2022matryoshka}. In our approach, the teacher-dimensional projection is used only during distillation and removed afterwards. Since direct vector alignment requires equal dimensionality, \(L_1\) is applied only to the projected student embeddings. By contrast, \(L_2\) and \(L_3\) operate on pairwise similarities and can therefore compare representation spaces with different dimensionalities. We apply these two losses to both the native and projected student embeddings. This allows the native student representation to preserve the distilled similarity structure after the projection layer is removed, reducing the number of parameters and inference cost while improving compatibility with embedding deployment frameworks that do not support custom projection layers.

\item \textbf{Multilingual training -}
While the original method was applied in a monolingual setting, we extend it to multilingual data. For each training example, one language is sampled randomly from the translations available in the corpus. Consequently, individual batches contain texts in multiple languages, and the relational distillation losses are optimized jointly across their representations.
\end{itemize}

\subsection{Contrastive fine-tuning}

The final stage optimizes the models directly for retrieval using the InfoNCE objective \citep{chen2020simple}. Each training instance is a triplet consisting of a query, a positive passage, and a hard negative. When multiple positives or hard negatives are available for a query, one passage from each group is sampled randomly. The negative set for each query includes both its assigned hard negative and the passages associated with other queries in the batch, which serve as in-batch negatives. All models are trained using the same hyperparameters: a batch size of 1,024 triplets, a temperature of 0.01, and 10 training epochs. The learning rate is linearly increased for 200 warm-up steps to a maximum value of \(2 \times 10^{-6}\), followed by linear decay.

A distinctive aspect of this stage is the construction of positive and negative examples. Although the training data are derived from publicly available retrieval datasets, we do not use their original relevance labels. Instead, candidate passages are relabeled using BGE-Reranker-v2.5-Gemma2-Lightweight\footnote{\url{https://huggingface.co/BAAI/bge-reranker-v2.5-gemma2-lightweight}}. For each query, we construct a candidate pool by retrieving the top 16 passages independently with a dense retriever, SPLADE, and BM25. Merging and deduplicating the results produces at most 48 candidate passages per query. Each query-passage pair is subsequently evaluated by the reranker. Using the raw scores produced by the reranker, passages scoring above 26 are selected as positives, whereas those scoring between 12 and 20 are treated as hard negatives, all remaining candidates are discarded. These thresholds were selected empirically in preliminary contrastive fine-tuning experiments. Models trained using this selection process consistently outperformed those trained using the original ground-truth relevance labels.

\begin{figure*}
  \centering
  \includegraphics[scale=0.48]{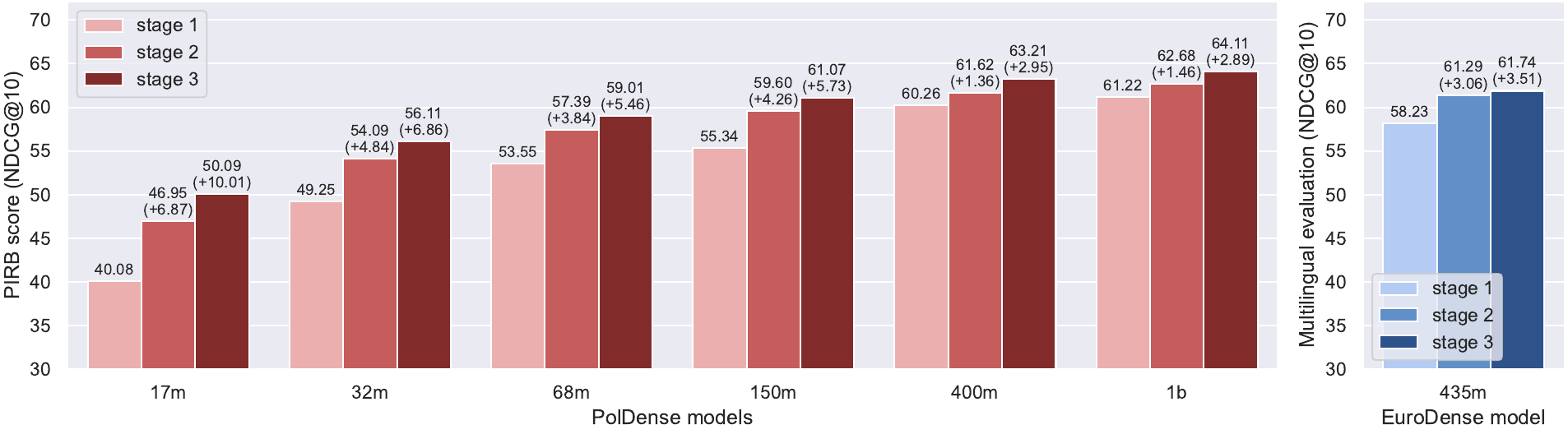}
  \caption{Retrieval performance after each stage of the training pipeline for PolDense and EuroDense models.}
  \label{fig:stages}
\end{figure*}

\subsection{Polish models (PolDense)}

We trained six Polish dense retrieval models ranging from 17M to 1B parameters. The models were initialized from the Ettin encoder family \citep{weller2026seq}, which served as the students during distillation. We used BGE-Multilingual-Gemma2 \citep{chen2024m3} as the teacher in both distillation stages. At the time it was the highest-performing model with a permissive license on the Polish Information Retrieval Benchmark (PIRB) \citep{dadas2024pirb}. The first-stage corpus comprised approximately 20 million English-Polish parallel text pairs. It was constructed primarily from English corpora that we translated into Polish using Gemma~3~27B \citep{team2025gemma}. The models were trained for five epochs with a batch size of 64, 1,000 warm-up steps, a maximum learning rate of \(2 \times 10^{-5}\), and linear learning-rate decay. Parallel data were used only in this stage, all subsequent training was conducted exclusively on Polish texts.

For the second stage, we augmented the Polish side of the parallel corpus with documents from the Polish subset of \textit{FineTranslations} \citep{penedo2026finetranslations}, selecting those with the highest \texttt{edu\_score} values. We added 50 million documents for the 17M-150M models and 13 million for the 400M and 1B models, resulting in corpora of approximately 70 and 33 million texts, respectively. The models were trained for five epochs with a batch size of 128, the gradual-unfreezing schedule described above, and a learning rate of \(1 \times 10^{-4}\).

The contrastive fine-tuning stage used 13 retrieval datasets comprising more than 4.5 million queries and 15 million passages. We applied the common fine-tuning configuration described in the preceding section. Detailed information about the datasets used at each training stage is provided in Appendix~\ref{app:training_data}.

\subsection{Multilingual model (EuroDense)}

We additionally trained EuroDense, a 435M-parameter model supporting nine European languages: English, German, French, Spanish, Italian, Portuguese, Dutch, Russian, and Polish. For both distillation stages we used a corpus of 20 million texts translated into all nine languages using Gemma~3~27B \citep{team2025gemma}. This produced approximately 180 million texts. Training was initialized from Stella-400M \citep{zhang2024jasper}, a strong English embedding model rather than a general-purpose pretrained encoder. In the first stage, we performed self-distillation using a frozen copy of Stella-400M as the teacher and another copy, initialized with the same weights, as the student. This extended the original embedding space to eight additional languages. Training lasted five epochs with a batch size of 64, 1,000 warm-up steps, a maximum learning rate of \(2 \times 10^{-5}\), and linear learning-rate decay.

For the second stage, we evaluated Qwen3-Embedding-8B \citep{zhang2025qwen3}, INF-Retriever-v1 \citep{infly-ai_2025}, QZhou-Embedding \citep{yu2025qzhouembeddingtechnicalreport}, and Pplx-Embed-v1-4B \citep{eslami2026diffusion} as potential teachers. Pplx-Embed achieved the strongest results in our preliminary distillation experiments, which are descibed in Appendix \ref{app:teacher_selection}, and was therefore selected. The student was trained for five epochs with a batch size of 128, the gradual-unfreezing schedule described above, and a maximum learning rate of \(1 \times 10^{-5}\). We used a lower learning rate than for polish models because this type of distillation was less stable at higher learning rates in the multilingual setting.

The final contrastive fine-tuning stage used the same hyperparameters as for PolDense and 11 retrieval datasets comprising approximately 1.6 million queries and more than 13 million passages. All datasets were machine-translated into the nine supported languages. During training, the language of each triplet component (query, positive, negative) was sampled randomly from its available translations. Further details on the datasets used at each stage are provided in Appendix~\ref{app:training_data}.

\section{Evaluation}
To assess the quality of PolDense and EuroDense, we conducted an extensive evaluation across a large collection of retrieval tasks and compared the models with popular multilingual and monolingual embedding models. The Polish models were evaluated on the PIRB benchmark \citep{dadas2024pirb}, which comprises 41 retrieval tasks. PIRB consolidates existing Polish retrieval evaluation suites and complements them with 10 additional datasets. For the multilingual evaluation, we assembled a broad collection of tasks covering all nine supported languages across diverse dataset domains and structures. The collection draws on established evaluation suites, including MTEB \citep{muennighoff2023mteb}, MMTEB \citep{enevoldsen2025mmteb}, RTEB \citep{aarsen2025rteb}, PIRB \citep{dadas2024pirb}, MTEB-French \citep{ciancone2024mtebfrenchresourcesfrenchsentence}, MTEB-NL \citep{banar-etal-2026-mteb}, BEIR-NL \citep{lotfi-etal-2025-beir}, and RusBEIR \citep{kovalev2025buildingrussianbenchmarkevaluation}, and is supplemented with numerous standalone retrieval tasks for the respective languages. In total, the multilingual evaluation comprised 150 datasets. For more information on the datasets used, see Appendix~\ref{app:evaluation_data}.

\begin{table}[t]
\centering
\scriptsize
\renewcommand{\arraystretch}{1.1}
\begin{tabular}{m{4cm}|c}
\hline
 \textbf{Model name} & \makecell[c]{\bf PIRB \\ \bf (NDCG@10)} \\
\hline
 \multicolumn{2}{l}{\textbf{Bigger models ($\ge$ 1B parameters)}} \\
\hline
 Qwen3-Embedding-4B               & 59.19 \\
 Qwen3-Embedding-8B               & 60.93 \\
 Inf-Retriever-v1                 & 61.52 \\
 Pplx-Embed-v1-4B                 & 62.09 \\
 Stella-PL-Retrieval-8K           & \cellcolor{bronze}{62.69} \\
 BGE-Multilingual-Gemma2          & \cellcolor{silver}{63.26} \\
 \textbf{PolDense-1B}             & \cellcolor{gold}{64.11} \\
\hline
 \multicolumn{2}{l}{\textbf{Smaller models ($<$ 1B parameters)}} \\
\hline
 BM25                             & 45.71 \\
 \textbf{PolDense-17M}            & 50.09 \\
 Multilingual-E5-Small            & 50.65 \\
 Qwen3-Embedding-0.6B             & 50.71 \\
 Multilingual-E5-Base             & 53.12 \\
 Silver-Retriever-Base-v1         & 53.33 \\
 Voyage-4-Nano                    & 55.15 \\
 Embeddinggemma-300M              & 55.22 \\
 BGE-M3                           & 55.75 \\
 \textbf{PolDense-32M}            & 56.11 \\
 Harrier-OSS-v1-0.6B              & 56.45 \\
 Jina-Embeddings-v5-Text-Small    & 57.19 \\
 Multilingual-E5-Large            & 57.29 \\
 Jina-Embeddings-v3               & 57.33 \\
 \textbf{PolDense-68M}            & 59.01 \\
 Pplx-Embed-v1-0.6B               & 59.13 \\
 Snowflake-Arctic-Embed-L-v2.0    & 59.22 \\
 \textbf{PolDense-150M}           & \cellcolor{bronze}{61.07} \\
 Stella-PL-Retrieval-Mini-8K      & \cellcolor{silver}{61.29} \\
 \textbf{PolDense-400M}           & \cellcolor{gold}{63.21} \\
\hline
\end{tabular}
\caption{Evaluation results on 41 Polish retrieval tasks.}
\label{tab:poldense_results}
\end{table}

In addition to our models, the evaluation included the Multilingual-E5 \citep{wang2024multilingual}, Qwen3-Embedding \citep{zhang2025qwen3}, Pplx-Embed \citep{eslami2026diffusion}, Jina-Embeddings \citep{akram2026jina}, Snowflake-Arctic-Embed \citep{yu2024arctic}, and BGE \citep{chen2024m3} model families. We additionally evaluated the multilingual EmbeddingGemma-300M \citep{vera2025embeddinggemma}, Voyage-4-Nano \citep{voyage_ai_2026}, Harrier-OSS-v1-0.6B\footnote{\url{https://hf.co/microsoft/harrier-oss-v1-0.6b}}, GTE-Multilingual-Base \citep{zhang2024mgte}, and INF-Retriever-v1 \citep{infly-ai_2025} models. For the Polish evaluation, the comparison was further extended with the Polish-specific Stella-PL-Retrieval \citep{dadas2024pirb} models and Silver-Retriever-Base-v1 \citep{rybak2024silver}. Even more models are included in Appendix \ref{app:monolingual_eval}, where we discuss the evaluation results broken down by language.

Figure~\ref{fig:stages} compares the performance of PolDense and EuroDense after each of the three training stages. Each stage improves retrieval quality, with the overall gains tending to be larger for smaller models. Contrastive fine-tuning provides gains ranging from 1.43 to 3.14 NDCG@10 points for PolDense, whereas the improvement for EuroDense is only 0.45 points on the multilingual evaluation suite. This smaller gain may partly reflect the \emph{curse of multilinguality} \citep{conneau2020unsupervised}: with fixed model capacity, supporting multiple languages introduces a trade-off between cross-lingual transfer and language-specific performance.

\begin{figure}
  \centering
  \includegraphics[scale=0.48]{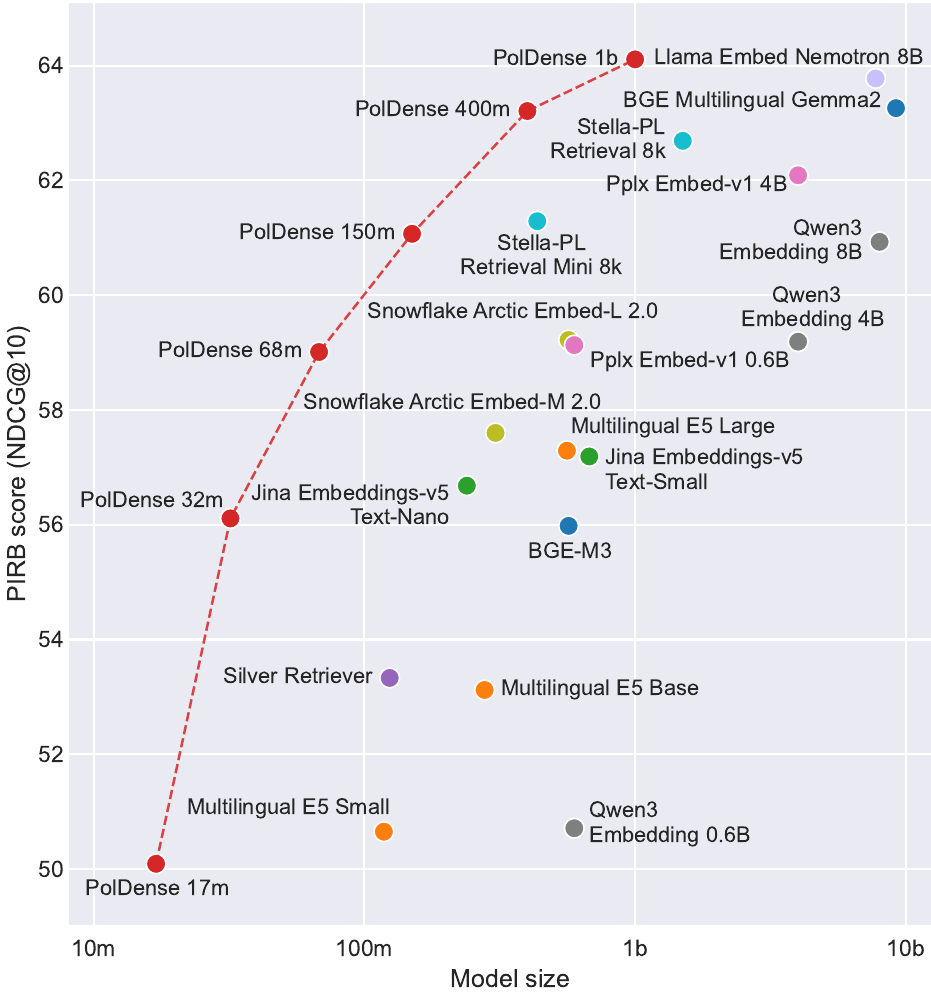}
  \caption{Retrieval quality as a function of model size on the PIRB benchmark. Each point reports mean NDCG@10. Points farther toward the upper left represent a more favorable quality-size trade-off.}
  \label{fig:poldense_frontier}
\end{figure}

\begin{table*}[!ht]
\scriptsize
\centering
\renewcommand{\arraystretch}{1.1}
\begin{tabular}{l|rrrrrrrrr|cc}
\hline
 \makecell[l]{\bf Model name \\ \bf  / (\# tasks)} &  
 \makecell[c]{\bf PL \\ \bf (41)} &  
 \makecell[c]{\bf DE \\ \bf (6)} &    
 \makecell[c]{\bf FR \\ \bf (9)} &    
 \makecell[c]{\bf ES \\ \bf (6)} &    
 \makecell[c]{\bf IT \\ \bf (7)} &
 \makecell[c]{\bf PT \\ \bf (8)} &
 \makecell[c]{\bf NL \\ \bf (25)} &    
 \makecell[c]{\bf RU \\ \bf (26)} & 
 \makecell[c]{\bf EN \\ \bf (22)} &
 \makecell[c]{\bf Avg. \\ \bf (150 tasks)} &   
 \makecell[c]{\bf Avg. \\ \bf (9 langs)} \\
 \hline
 \multicolumn{11}{l}{\textbf{Bigger models ($\ge$ 1B parameters)}} \\
 \hline
 Qwen3-Embedding-4B      & 59.19                     & \cellcolor{bronze}{68.16} & 64.01                     & \cellcolor{bronze}{66.98} & 72.05                     & 67.93                     & 58.68                     & 62.88                     & \cellcolor{bronze}{63.57} & 62.41                     & 64.83                     \\
 Qwen3-Embedding-8B      & 60.93                     & 67.34                     & 64.38                     & 66.95                     & 72.59                     & 69.59                     & \cellcolor{gold}{60.52}   & \cellcolor{silver}{64.64} & \cellcolor{gold}{65.56}   & \cellcolor{bronze}{63.89} & 65.83                     \\
 Pplx-Embed-v1-4b        & \cellcolor{silver}{62.09} & \cellcolor{gold}{70.76}   & \cellcolor{silver}{66.19} & 66.16                     & \cellcolor{bronze}{73.11} & \cellcolor{bronze}{69.85} & 59.40                      & 64.06                     & 62.31                     & 63.70                      & \cellcolor{bronze}{65.99} \\
 BGE-Multilingual-Gemma2 & \cellcolor{gold}{63.26}   & 66.76                     & \cellcolor{bronze}{65.40}  & \cellcolor{gold}{67.39}   & \cellcolor{gold}{74.54}   & \cellcolor{gold}{72.10}    & \cellcolor{bronze}{59.74} & \cellcolor{bronze}{64.59} & 60.39                     & \cellcolor{silver}{63.92} & \cellcolor{silver}{66.02} \\
 Inf-Retriever-v1         & \cellcolor{bronze}{61.52} & \cellcolor{silver}{68.87} & \cellcolor{gold}{66.65}   & \cellcolor{silver}{67.18} & \cellcolor{silver}{74.36} & \cellcolor{silver}{71.44} & \cellcolor{silver}{60.02} & \cellcolor{gold}{64.94}   & \cellcolor{silver}{63.93} & \cellcolor{gold}{64.17}   & \cellcolor{gold}{66.55}   \\
\hline
 \multicolumn{11}{l}{\textbf{Smaller models ($<$ 1B parameters)}} \\
 \hline
  Multilingual-E5-Small         & 50.65                     & 55.24                     & 45.47                     & 52.31                     & 63.06                     & 56.82                     & 46.21                     & 52.13                     & 45.89                     & 50.32                     & 51.98                     \\
 Multilingual-E5-Base          & 53.12                     & 56.64                     & 50.37                     & 54.17                     & 64.84                     & 59.67                     & 48.07                     & 54.24                     & 48.20                      & 52.67                     & 54.37                     \\
 GTE-Multilingual-Base         & 51.61                     & 60.59                     & 55.79                     & 58.91                     & 65.47                     & 61.01                     & 48.75                     & 55.15                     & 51.94                     & 53.85                     & 56.58                     \\
 Multilingual-E5-Large         & 57.29                     & 56.85                     & 52.67                     & 57.21                     & 66.72                     & 63.05                     & 51.63                     & 57.70                      & 51.27                     & 55.99                     & 57.15                     \\
 Qwen3-Embedding-0.6B          & 50.71                     & 61.66                     & 56.99                     & 59.07                     & 63.98                     & 62.32                     & 49.99                     & 56.03                     & 56.99                     & 54.82                     & 57.53                     \\
 BGE-M3                        & 55.98                     & 63.13                     & 57.53                     & 60.01                     & 67.98                     & 65.91                     & 51.16                     & 58.03                     & 50.36                     & 56.34                     & 58.90                      \\
 Jina-Embeddings-v3            & 57.33                     & 63.35                     & 57.97                     & 59.20                      & 66.04                     & 63.91                     & 53.82                     & 58.01                     & 53.58                     & 57.42                     & 59.25                     \\
 Snowflake-Arctic-Embed-M-v2.0 & 57.60                      & 62.79                     & 59.43                     & 60.04                     & 68.21                     & 66.59                     & 49.85                     & 54.26                     & 55.34                     & 56.79                     & 59.35                     \\
 Embeddinggemma-300M           & 55.22                     & 64.40                      & 60.76                     & 60.99                     & 68.07                     & 66.47                     & 53.28                     & 60.07                     & 55.07                     & 57.85                     & 60.48                     \\
 Harrier-OSS-v1-0.6B           & 56.45                     & 64.88                     & 60.16                     & 60.33                     & 68.20                      & 64.33                     & 54.15                     & 59.77                     & 57.19                     & 58.43                     & 60.61                     \\
 Jina-Embeddings-v5-Text-Nano  & 56.68                     & 64.71                     & 60.91                     & \cellcolor{bronze}{63.22} & 69.10                      & 67.40                     & 55.40                      & 60.14                     & 57.69                     & 59.20                      & 61.69                     \\
 Voyage-4-Nano                 & 55.15                     & \cellcolor{silver}{68.81} & \cellcolor{silver}{63.12} & \cellcolor{silver}{63.83} & \cellcolor{silver}{70.49} & \cellcolor{bronze}{67.61} & 52.31                     & 58.51                     & 55.91                     & 58.12                     & 61.75                     \\
 Snowflake-Arctic-Embed-L-v2.0 & \cellcolor{silver}{59.22} & \cellcolor{bronze}{67.31} & 60.30                      & 61.60                      & 69.46                     & \cellcolor{gold}{68.13}   & 55.73                     & 59.78                     & 54.90                      & 59.54                     & 61.83                     \\
 Jina-Embeddings-v5-Text-Small & 57.19                     & 65.18                     & 61.08                     & 61.85                     & 68.93                     & 66.70                      & \cellcolor{bronze}{56.22} & \cellcolor{bronze}{60.88} & \cellcolor{bronze}{58.65} & \cellcolor{bronze}{59.68} & \cellcolor{bronze}{61.85} \\
 Pplx-Embed-v1-0.6B            & \cellcolor{bronze}{59.13} & 66.87                     & \cellcolor{bronze}{62.01} & 63.01                     & \cellcolor{bronze}{69.73} & \cellcolor{silver}{68.06} & \cellcolor{silver}{56.88} & \cellcolor{silver}{61.31} & \cellcolor{gold}{59.87}   & \cellcolor{silver}{60.85} & \cellcolor{silver}{62.99} \\
 \bf{EuroDense-435M}          & \cellcolor{gold}{61.16}   & \cellcolor{gold}{68.94}   & \cellcolor{gold}{63.36}   & \cellcolor{gold}{64.57}   & \cellcolor{gold}{71.43}   & 66.41                     & \cellcolor{gold}{58.11}   & \cellcolor{gold}{61.36}   & \cellcolor{silver}{59.22} & \cellcolor{gold}{61.74}   & \cellcolor{gold}{63.84}   \\
\hline
\end{tabular}
\caption{\label{tab:eurodense_results}
Results of multilingual models across nine languages and 150 retrieval tasks (NDCG@10).
}
\end{table*}

Table~\ref{tab:poldense_results} compares PolDense with Polish and multilingual embedding models on PIRB. The models are divided into two groups: those with fewer than 1B parameters and those with at least 1B parameters. Within each group, the three highest scores are highlighted in gold, silver, and bronze. PolDense achieves a particularly strong trade-off between retrieval quality and model size. Among the larger models, PolDense-1B obtains the highest NDCG@10 score, outperforming substantially larger models such as BGE-Multilingual-Gemma2 (9B parameters), INF-Retriever-v1 (7B), and Qwen3-Embedding-8B. In the smaller-model group, PolDense-400M and PolDense-150M rank first and third, respectively. Even the compact PolDense-32M and PolDense-68M variants remain competitive with multilingual models containing several hundred million parameters. Figure~\ref{fig:poldense_frontier} illustrates this efficiency by relating PIRB performance to model size. PolDense variants form the Pareto frontier across evaluated sizes, with retrieval quality consistently improving as capacity increases. Multiple sizes enable practical deployment choices: smaller variants suit resource-constrained, low-latency environments, whereas larger ones prioritize retrieval quality.

Table~\ref{tab:eurodense_results} presents the multilingual evaluation results. We report mean NDCG@10 by language and two aggregates: the mean over 150 tasks and a language-balanced mean. Models are grouped by parameter count, with the three best results per group highlighted. In the small models group, EuroDense-435M achieves the best result in seven of the nine languages and ranks first according to both aggregate measures. Pplx-Embed-v1-0.6B and Jina-Embeddings-v5-Text-Small rank second and third, trailing EuroDense by approximately one and two NDCG@10 points, respectively. Voyage-4-Nano also performs strongly for several languages, but its results are less consistent across the complete evaluation suite. EuroDense therefore achieves the best overall performance in its size class, although its parameter efficiency relative to multi-billion-parameter models is less pronounced than that of PolDense. It remains competitive with the larger models for individual languages, particularly Polish and German, but performs below them on both aggregate measures. Detailed results broken down by individual tasks are shown in Appendix \ref{app:multilingual_eval}.

\section{Conclusions}
We introduced PolDense and EuroDense, parameter-efficient dense retrievers trained through cross-lingual alignment, relational distillation, and contrastive fine-tuning. Across 41 Polish tasks, PolDense provides a strong quality-size trade-off, with PolDense-1B establishing a new state of the art on PIRB. EuroDense-435M achieves the best aggregate performance among the evaluated multilingual models below one billion parameters, leading in seven of nine languages. These results demonstrate that carefully designed distillation and data relabeling can produce compact retrievers competitive with substantially larger models. By releasing all models publicly, we aim to facilitate their adoption in practical, multilingual retrieval and RAG systems.

\section*{Limitations}

\paragraph{Retrieval-oriented training and evaluation.}
This work focuses specifically on document retrieval in natural languages, and both the training pipeline and evaluation were designed for this setting. Many of the models used in our comparison are general-purpose embeddings that also support tasks such as semantic textual similarity, clustering, and few-shot classification. PolDense and EuroDense were not explicitly optimized or evaluated for these tasks and may therefore underperform more general-purpose models outside retrieval. Accordingly, the reported results should be interpreted as evidence of retrieval quality and parameter efficiency rather than overall embedding versatility.

\paragraph{Computational constraints.}
Limited computational resources restricted the scope of some experiments. To reduce training cost, the 400M and 1B PolDense models used only 13 million documents from \textit{FineTranslations} \citep{penedo2026finetranslations}, compared with 50 million for the smaller variants. Consequently, comparisons across model sizes reflect differences in both model capacity and the amount of distillation data. Multilingual data preparation and training were substantially more expensive, we therefore trained only one EuroDense variant with 435M parameters, and limited its coverage to nine languages. Languages were selected from among European languages based primarily on their estimated number of speakers worldwide, except for Polish, which was included because the required data had already been available. Our experiments therefore do not examine multilingual scaling across model sizes or generalization to other languages.

\section*{Ethical considerations}

\paragraph{Use of AI tools.}
AI tools were used in the preparation of the publication for only language editing, grammatical correction and minor stylistic improvements. They are not used for generating the core scientific content.

\section*{Acknowledgments}
The research was supported by the project Large Language Models for the European Union (LLMs4EU). This project is co-funded by the Digital Europe Programme under Grant Agreement 101198470.

The research was supported [in part] by project “Cloud Artificial Intelligence Service Engineering (CAISE) platform to create universal and smart services for various application areas”, No. KPOD.05.10-IW.10-0005/24, as part of the European IPCEI-CIS program, financed by NRRP (National Recovery and Resilience Plan) funds. Computations were carried out using the computers of Centre of Informatics Tricity Academic Supercomputer \& Network at Gdansk University of Technology.

\bibliography{custom}

\clearpage
\appendix

\renewcommand*\footnoterule{} 
\newcommand\blfootnote[1]{
  \begingroup
  \renewcommand\thefootnote{}\footnote{#1}
  \addtocounter{footnote}{-1}
  \endgroup
}

\section{Training data}
\label{app:training_data}

This section summarizes the datasets used to train PolDense and EuroDense. Table~\ref{tab:distillation_data} lists the corpora used in the first two training stages. The corpus is divided into two groups: question-like and passage-like texts. For each dataset, we report its name, source, and number of records. The \textbf{PD} and \textbf{ED} columns indicate whether it was used to train PolDense or EuroDense, respectively. All listed datasets except FineTranslations \citep{penedo2026finetranslations} were used in both distillation stages for both model families. For FineTranslations, we used only the Polish subset to augment the second-stage corpus of PolDense. Most corpora were obtained from external sources, while the following three were prepared internally:

\noindent \textbf{(1) ELI5 (generated)} consists of synthetic answers generated for questions from the original ELI5 dataset using Gemma~3~27B \citep{team2025gemma}. The answers were subsequently translated into all supported languages using the same model. This corpus is our extension of ELI5 and is not part of the original dataset.

\noindent  \textbf{(2) Sci-Abstract} contains abstracts of scientific publications and research projects. The initial corpus comprised more than 80,000 abstracts in Polish and English. For EuroDense, the abstracts were machine-translated into the remaining supported languages.

\noindent \textbf{(3) Wikipedia} was initially constructed as a parallel corpus of more than 40,000 paragraphs mined from the Polish and English Wikipedia using bitext-mining methods. For EuroDense, the paragraphs were machine-translated into the remaining supported languages.

Table~\ref{tab:fine_tuning_data} lists the retrieval datasets used during the fine-tuning stage. Each dataset contains queries and passages, whose respective counts are reported in the \textbf{Queries} and \textbf{Passages} columns. The table additionally provides the dataset name and source, while the \textbf{PD} and \textbf{ED} columns indicate whether it was included in the fine-tuning data for PolDense or EuroDense.

\begin{table}[t]
\centering
\small
\begin{tabular}{lrrr}
\toprule
\textbf{Teacher} &
\textbf{\shortstack{Teacher\\score}} &
\textbf{\shortstack{Student\\score}} &
\textbf{Diff.} \\
\midrule
Qwen3-Embedding-8B & 71.0 & 64.0 & 7.0 \\
INF-Retriever-v1   & 72.9 & 61.6 & 11.3 \\
QZhou-Embedding    & 72.3 & 62.5 & 9.8 \\
Pplx-Embed-v1-4B   & 69.5 & \textbf{64.5} & \textbf{5.0} \\
\bottomrule
\end{tabular}
\caption{Teacher-selection results on NanoBEIR. \emph{Diff.} denotes the difference between the teacher and distilled student scores. Values are mean NDCG@10 across the benchmark tasks.}
\label{tab:teacher_comparison}
\end{table}

\begin{table*}[!ht]
\centering
\small
\setlength{\tabcolsep}{8pt}
\renewcommand{\arraystretch}{0.90}
\begin{tabular}{llrcc}
\toprule
Dataset & Reference & Records & PD & ED \\
\midrule
\multicolumn{5}{l}{\textbf{Query corpus}} \\
\midrule
ELI5            & \citep{fan-etal-2019-eli5}              & 508,699   & $\checkmark$ & $\checkmark$ \\
FiQA             & \citep{maia-etal-2018-fiqa}             & 6,646     & $\checkmark$ & $\checkmark$ \\
GooAQ            & \citep{khashabi-etal-2021-gooaq}        & 2,998,852 & $\checkmark$ & $\checkmark$ \\
HotpotQA         & \citep{yang-etal-2018-hotpotqa}         & 97,851    & $\checkmark$ & $\checkmark$ \\
LFQA             & \citep{krishna-etal-2021-hurdles}       & 229,166   & $\checkmark$ & $\checkmark$ \\
MS MARCO         & \citep{bajaj-etal-2016-msmarco}         & 808,731   & $\checkmark$ & $\checkmark$ \\
Natural Questions & \citep{kwiatkowski-etal-2019-natural}  & 106,921   & $\checkmark$ & $\checkmark$ \\
SciDocs          & \citep{cohan-etal-2020-specter}         & 208,248   & $\checkmark$ & $\checkmark$ \\
SciFact          & \citep{wadden-etal-2020-scifact}        & 1,109     & $\checkmark$ & $\checkmark$ \\
WikiHow          & \citep{koupaee-wang-2018-wikihow}       & 128,333   & $\checkmark$ & $\checkmark$ \\
ZnanyLekarz      & \citep{dadas2024assessing}              & 76,324    & $\checkmark$ & $\checkmark$ \\
\midrule
\multicolumn{5}{l}{\textbf{Passage corpus}} \\
\midrule
ALLNLI           & \citep{bowman-etal-2015-snli,williams-etal-2018-multinli} & 783,963 & $\checkmark$ & $\checkmark$ \\
ELI5             & \citep{fan-etal-2019-eli5}              & 1,110,547 & $\checkmark$ & $\checkmark$ \\
ELI5 (generated) & See Appendix~\ref{app:training_data} (1) & 508,623   & $\checkmark$ & $\checkmark$ \\
FiQA             & \citep{maia-etal-2018-fiqa}             & 17,070    & $\checkmark$ & $\checkmark$ \\
GooAQ            & \citep{khashabi-etal-2021-gooaq}        & 1,471,467 & $\checkmark$ & $\checkmark$ \\
LingNLI          & \citep{parrish-etal-2021-linguist}       & 34,910    & $\checkmark$ & $\checkmark$ \\
Quora            & \citep{iyer-etal-2017-quora}            & 147,776   & $\checkmark$ & $\checkmark$ \\
PolGov           & \citep{dadas2024assessing}              & 10,976    & $\checkmark$ & $\checkmark$ \\
LFQA             & \citep{krishna-etal-2021-hurdles}       & 228,808   & $\checkmark$ & $\checkmark$ \\
MS MARCO         & \citep{bajaj-etal-2016-msmarco}         & 8,841,823 & $\checkmark$ & $\checkmark$ \\
Sci-Abstracts & See Appendix~\ref{app:training_data} (2)   & 83,986    & $\checkmark$ & $\checkmark$ \\
PELCRA           & \citep{pezik2018spokesmix}               & 15,905    & $\checkmark$ & $\checkmark$ \\
SciDocs          & \citep{cohan-etal-2020-specter}         & 207,490   & $\checkmark$ & $\checkmark$ \\
SciFact          & \citep{wadden-etal-2020-scifact}        & 5,183     & $\checkmark$ & $\checkmark$ \\
Tatoeba          & \citep{artetxe-schwenk-2019-massively}  & 78,002    & $\checkmark$ & $\checkmark$ \\
Wikipedia    &  See Appendix~\ref{app:training_data} (3)   & 46,340    & $\checkmark$ & $\checkmark$ \\
WikiHow          & \citep{koupaee-wang-2018-wikihow}       & 128,177   & $\checkmark$ & $\checkmark$ \\
ZnanyLekarz      & \citep{dadas2024assessing}              & 123,790   & $\checkmark$ & $\checkmark$ \\
FineTranslations & \citep{penedo2026finetranslations}      & 13M / 50M & $\checkmark$ &              \\
\bottomrule
\end{tabular}
\caption{Summary of the datasets used for training in the first two stages (distillation). PD and ED refer to the PolDense and EuroDense models, respectively.}
\label{tab:distillation_data}
\end{table*}

\begin{table*}[!ht]
\centering
\small
\setlength{\tabcolsep}{9pt}
\renewcommand{\arraystretch}{0.90}
\begin{tabular}{llrrcc}
\toprule
Dataset & Reference & Queries & Passages & PD & ED \\
\midrule
ArguAna      & \citep{wachsmuth-etal-2018-counterargument} & 3,101     & 6,786     & $\checkmark$ & $\checkmark$ \\
ELI5         & \citep{fan-etal-2019-eli5}                  & 508,699   & 1,110,547 & $\checkmark$ & $\checkmark$ \\
FEVER        & \citep{thorne-etal-2018-fever}              & 102,292   & 816,468   & $\checkmark$ & $\checkmark$ \\
FiQA         & \citep{maia-etal-2018-fiqa}                 & 5,500     & 39,118    & $\checkmark$ & $\checkmark$ \\
GooAQ        & \citep{khashabi-etal-2021-gooaq}            & 2,998,852 & 1,471,467 & $\checkmark$ &              \\
PolGov       & \citep{dadas2024assessing}                  & 69,888    & 38,191    & $\checkmark$ &              \\
HotpotQA     & \citep{yang-etal-2018-hotpotqa}             & 84,994    & 1,514,002 & $\checkmark$ & $\checkmark$ \\
MS MARCO     & \citep{bajaj-etal-2016-msmarco}             & 808,731   & 8,841,823 & $\checkmark$ & $\checkmark$ \\
NFCorpus     & \citep{boteva-etal-2016-nfcorpus}           & 2,576     & 3,633     & $\checkmark$ & $\checkmark$ \\
NQ           & \citep{kwiatkowski-etal-2019-natural}       & 58,564    & 1,117,418 & $\checkmark$ & $\checkmark$ \\
SciDocs      & \citep{cohan-etal-2020-specter}             & 884       & 17,423    & $\checkmark$ & $\checkmark$ \\
SciFact      & \citep{wadden-etal-2020-scifact}            & 807       & 4,961     & $\checkmark$ & $\checkmark$ \\
ZnanyLekarz  & \citep{dadas2024assessing}                                             & 76,324    & 123,790   & $\checkmark$ & $\checkmark$ \\
\bottomrule
\end{tabular}
\caption{Summary of the datasets used for fine-tuning. PD and ED refer to the PolDense and EuroDense models, respectively.}
\label{tab:fine_tuning_data}
\end{table*}

\section{Teacher selection}
\label{app:teacher_selection}

We conducted a controlled experiment to examine how the choice of teacher affects student performance during relational distillation. The experiment was restricted to English and used the English portion of the distillation corpus. The student was initialized from mmBERT-base \citep{marone2025mmbert}, while Qwen3-Embedding-8B \citep{zhang2025qwen3}, INF-Retriever-v1 \citep{infly-ai_2025}, QZhou-Embedding \citep{yu2025qzhouembeddingtechnicalreport}, and Pplx-Embed-v1-4B \citep{eslami2026diffusion} were evaluated as teachers. For each teacher, we performed the complete relational-distillation stage. The student was trained for five epochs with a batch size of 128, the gradual-unfreezing schedule described, and a peak learning rate of \(1 \times 10^{-4}\). During training, performance was monitored on the English NanoBEIR benchmark, a compact 13-task subset of BEIR designed for computationally efficient evaluation.\footnote{\url{https://docs.mteb.org/overview/available_benchmarks/\#nanobeir}}

As shown in Table~\ref{tab:teacher_comparison}, a stronger teacher does not necessarily produce a stronger student. Pplx-Embed-v1-4B obtained the lowest teacher score but yielded the best student result and the smallest teacher-student gap. Qwen3-Embedding-8B produced a slightly weaker student, while distillation from INF-Retriever-v1 and QZhou-Embedding resulted in substantially lower scores. These results suggest that standalone retrieval quality is not the only relevant criterion for teacher selection.

\section{Evaluation data}
\label{app:evaluation_data}

To compare the models we developed with existing ones, we conducted a comprehensive evaluation covering information retrieval tasks in various languages and domains. The tasks used were primarily drawn from monolingual benchmarks or from multilingual datasets such as MLRD \citep{chen2024m3}, MFQA \citep{de-bruyn-etal-2021-mfaq}, MKQA \citep{longpre-etal-2021-mkqa}, WebFAQ \citep{dinzinger2025webfaq}, or BelebeleRetrieval\footnote{\url{https://hf.co/datasets/mteb/belebele}} and WikipediaRetrievalMultilingual\footnote{\url{https://hf.co/datasets/mteb/WikipediaRetrievalMultilingual}} from MMTEB \citep{enevoldsen2025mmteb}. In cases where there were few tasks for a given language, we searched for additional publicly available datasets. The tasks have been adapted for execution within the PIRB framework\footnote{\url{https://github.com/sdadas/pirb}}, which is part of the benchmark for the Polish language. The exception was RTEB \citep{aarsen2025rteb} tasks, which were run using the MTEB framework\footnote{\url{https://github.com/embeddings-benchmark/mteb/}}. Details regarding the scope of tasks for each language are provided below. 

\paragraph{English}
The evaluation included 22 tasks from two benchmarks: BEIR \citep{thakur2021beir} and RTEB \citep{aarsen2025rteb}. BEIR is a well-established heterogeneous evaluation benchmark that covers various types of tasks, such as argument retrieval, duplicate detection, citation prediction, fact checking, entity retrieval, and domain-specific question answering, based on data from Wikipedia, scientific articles, and financial or medical documents. RTEB\footnote{\url{https://mteb-leaderboard.hf.space/benchmark/RTEB(eng, beta)}} is a recently proposed benchmark designed to reliably evaluate the retrieval accuracy of embedding models for real-world applications, including those in the financial and legal fields. In our evaluation, we used only publicly available tasks and omitted those related to code retrieval, as these fall outside the scope of our model's support.

\paragraph{German}

The German evaluation comprised six retrieval tasks: GerDaLIR \citep{wrzalik-krechel-2021-gerdalir}, which focuses on retrieving relevant German legal decisions; GermanDPR \citep{moller2021germanquad}, a passage-retrieval dataset for open-domain question answering; GermanGovServiceRetrieval \citep{lhm-dienstleistungen-qa}, which targets retrieval of public-service information from user questions; GermanQuAD-Retrieval \citep{moller2021germanquad}, which evaluates retrieval of answer-containing passages for German QA; LegalQuAD \citep{9723721}, a legal question-answering retrieval task; and the German subset of MLDR, which evaluates retrieval over long documents.  GerDaLIR, LegalQuAD, and MLDR additionally contain long documents, enabling evaluation across different input lengths and domains.

\paragraph{French}
The evaluation comprised nine tasks, five of which were French parts of MLDR, MFQA, MKQA, WebFAQ, and BelebeleRetrieval. Three tasks were taken from MTEB-French \citep{ciancone2024mtebfrenchresourcesfrenchsentence}, an extension of the English version of MTEB. MTEB-French provides French translations of existing tasks and introduces three new datasets, enabling a comprehensive evaluation across eight task categories. In our case, we selected only the retrieval tasks, that is, AlloprofRetrieval, SyntecRetrieval, and bBSARD. In addition, the CLIRudit \citep{valentini2025clirudit} dataset was included, which is an English–French academic retrieval dataset built from Érudit, a Canadian publishing platform.

\paragraph{Spanish}
The evaluation was conducted using the Spanish versions of MFQA, MKQA, and MLDR, as well as three Spanish-specific datasets. MessIRve \citep{valentini2025messirve} is a large-scale information retrieval dataset containing native queries collected through Google's Autocomplete API and paired with relevant documents sourced from Wikipedia, covering a broad range of topics. PRES \citep{Kamateri2019pres} is based on Spanish health-related web documents and covers 37 topics addressing general public health needs. SQAC \citep{Gutierrez2022MarIA} consists of question–answer pairs derived from articles from Wikinews, Spanish Wikipedia, and the AnCora corpus.

\paragraph{Italian}
The evaluation comprised seven tasks based on the Italian versions of MFQA, MLDR, WebFAQ, BelebeleRetrieval, and WikipediaRetrievalMultilingual, complemented by two datasets specifically designed for Italian. SQuAD-it \citep{Croce2018SQuAD} is a semi-automatically translated version of the English SQuAD dataset, designed for training and evaluating question-answering systems. WikipediaQA-ita \citep{wikipediaQA-ita} is an open-source Italian question-answering dataset containing synthetic question–context–answer pairs derived from Italian Wikipedia articles.

\paragraph{Portuguese}
The evaluation included five tasks based on Portuguese versions of MFQA, MKQA, MLDR, WebFAQ, and BelebeleRetrieval, along with three datasets created for the Portuguese language. JurisTCU \citep{juristcu2026} is a Legal Information Retrieval dataset in Brazilian Portuguese, consisting of jurisprudence from the Brazilian Federal Court of Accounts. Pirá \citep{pira2021} is a bilingual English–Portuguese QA dataset focusing on oceans, the Brazilian coast, and climate change. It also includes human-generated and automatically generated paraphrases of questions and answers. Quati \citep{bueno2024quati} is a dataset specifically designed for Brazilian Portuguese, consisting of questions posed by native speakers and a collection of documents from selected high-quality websites in Brazilian Portuguese.

\paragraph{Dutch}
The evaluation included 25 tasks drawn primarily from two Dutch benchmarks: BEIR-NL \citep{lotfi-etal-2025-beir} and MTEB-NL \citep{banar-etal-2026-mteb}.
BEIR-NL  is the Dutch version of BEIR, a zero-shot information retrieval benchmark. Fifteen datasets from BEIR-NL were used for the evaluation. MTEB-NL is a Dutch adaptation of the MTEB benchmark, comprising 12 datasets from the original MTEB, along with additional datasets that capture Dutch-specific language and domain characteristics. Five datasets from MTEB-NL were included in the evaluation. In addition, we used the Dutch versions of MFQA, MKQA, WebFAQ, BelebeleRetrieval, and WikipediaRetrievalMultilingual.

\paragraph{Russian}
The evaluation for the Russian language was based on the rusBEIR \citep{kovalev2025buildingrussianbenchmarkevaluation} benchmark and four tasks from the multilingual datasets MFQA, MKQA, MLDR, and BelebeleRetrieval. RusBEIR includes Russian translations of BEIR datasets, the Open-Source Dataset, and ruMTEB \citep{snegirev-etal-2025-russian}, as well as newly introduced datasets for fact-checking and information retrieval, based on data from Wikifacts. 

\paragraph{Polish}
We used the Polish Information Retrieval Benchmark (PIRB)  \cite{dadas2024pirb} to evaluate the models on Polish-language tasks. PIRB covers 41 Polish multidomain information retrieval tasks, including pre-existing datasets such as MaupQA \citep{rybak-2023-maupqa}, BEIR-PL \citep{wojtasik-etal-2024-beir}, and PolEval-2022 \citep{poleval_2022}, as well as a set of web datasets containing real questions and answers from selected Polish web services. In addition, PIRB includes the Polish version of the MFQA and the semi-automatically generated GPT-exams dataset.

\section{Detailed evaluation results}

The developed models were evaluated extensively, with an overview of the results provided in Tables \ref{tab:poldense_results} and \ref{tab:eurodense_results}. In the following subsections, we present detailed results showing a thorough comparison of our models with other multilingual and monolingual models. The list of models tested in this paper is provided in Table \ref{tab:models}.

\subsection{Multilingual evaluation results for individual tasks}
\label{app:multilingual_eval}

In Tables \ref{tab:detail_multilingaul_results} and \ref{tab:detail_multilingaul_results_2}, we have compiled detailed results for the top 10 smaller multilingual models with fewer than 1B parameters across all 150 tasks. Our EuroDense-435M model ranked among the top 3 models on 97 tasks (64.67\%) and took first place 62 times (41.33\%).
The EuroDense-435M model achieved its greatest advantage over the second model, based on the average across tasks, for the Polish language. This outcome was influenced by the fact that our model achieved the best results on nearly half of the tasks for this language, and that the best results for the remaining tasks were not attributed to a single model but were distributed among several models, notably Pplx-Embed-v1-0.6, Snowflake-Arctic-Embed-L-v2.0, and Voyage-4-Nano.
For German, the Voyage-4-Nano model achieved the best results across four tasks. Our model came in second in three of those cases and won two tasks, resulting in an average score across all German tasks slightly better than that of the Voyage-4-Nano model.
For French, the Voyage-4-Nano model also achieved a very good result compared to EuroDense-435M. Both models achieved the best results on three different tasks. As with German, the average score across all French tasks was higher for our model.
In Spanish, the best results for most tasks were achieved by different models. The exception is our model, which was the only one to achieve the best result on two tasks.
The highest scores for individual tasks in Italian were split among three models: Voyage-4-Nano, Pplx-Embed-v1-0.6B, and EuroDense-435M. The first two models each won two tasks, while our model won three, giving it an almost 1-percentage-point lead in the average score across all Italian tasks.
In Portuguese, the top results were mainly distributed among three models: Voyage-4-Nano, Snowflake-Arctic-Embed-L-v2.0, and Pplx-Embed-v1-0.6B. Our model performed worse than these models on most of the tasks designed specifically for Portuguese, but it achieved very good results on two Portuguese tasks from the multilingual datasets Belebele and MLDR, where it ranked first and second, respectively. 
In 60\% (15/25) of the tasks, the EuroDense-435M model achieved the best result for Dutch. The Pplx-Embed-v1-0.6B model came in second for this language, achieving the best result four times. Nevertheless, our model’s lead over the Pplx-Embed-v1-0.6B model across all tasks was 1.23 percentage points.
For Russian, the best models are EuroDense-435M and Pplx-Embed-v1-0.6B, which won 11 and 7 tasks, respectively. Despite this difference in the number of top results, the average score for the entire set of Russian tasks was similar for both models, and our model won by a narrow margin.
Nine of the ten models analyzed achieved the highest score on at least one English-language task. Our model had the highest number of top results (5). At the same time, its scores on some tasks were lower than those of the other models, so in the overall evaluation for this language, it ranked second, behind the Pplx-Embed-v1-0.6B model.

\begin{table*}[!ht]
\tiny
\centering
\renewcommand{\arraystretch}{1.1}
\begin{tabular}{lrrrrrrrrrr}
\hline
 Task                                        & \rotatebox{90}{Snowflake-Arctic-Embed-M-v2.0}   & \rotatebox{90}{Jina-Embeddings-v3}   & \rotatebox{90}{Embeddinggemma-300M}   & \rotatebox{90}{Voyage-4-Nano}   & \rotatebox{90}{Harrier-OSS-v1-0.6B}   & \rotatebox{90}{Jina-Embeddings-v5-Text-Nano}   & \rotatebox{90}{Snowflake-Arctic-Embed-L-v2.0}   & \rotatebox{90}{Jina-Embeddings-v5-Text-Small}   & \rotatebox{90}{Pplx-Embed-v1-0.6B}   & \rotatebox{90}{\bf{EuroDense-435M}} \\ 
\hline
 Gemini                                      & 97.91                                        & 98.28                             & \cellcolor{bronze}{98.59}             & 97.51                        & 96.44                              & 98.13                                       & \cellcolor{silver}{98.88}                       & 98.57                                        & 97.83                             & \cellcolor{gold}{98.91}          \\
 Odi                                         & 91.51                                        & \cellcolor{silver}{93.59}            & 91.51                              & \cellcolor{bronze}{93.56}       & 90.79                              & 90.51                                       & 92.94                                        & 90.80                                        & 92.15                             & \cellcolor{gold}{94.95}          \\
 Onet                                        & 57.95                                        & 57.45                             & 57.44                              & \cellcolor{gold}{62.92}         & 58.36                              & 56.64                                       & \cellcolor{bronze}{60.58}                       & 57.24                                        & 59.44                             & \cellcolor{silver}{61.54}        \\
\hline
\end{tabular}
\caption{\label{tab:detail_multilingaul_results}
Detailed evaluation results for ten multilingual models on tasks in Polish, German, French, Spanish, Italian, and Portuguese.
}
\end{table*}

\begin{table*}[!ht]
\tiny
\centering
\renewcommand{\arraystretch}{1.1}
\begin{tabular}{lrrrrrrrrrr}
\hline
 Task                                        & \rotatebox{90}{Snowflake-Arctic-Embed-M-v2.0}   & \rotatebox{90}{Jina-Embeddings-v3}   & \rotatebox{90}{Embeddinggemma-300M}   & \rotatebox{90}{Voyage-4-Nano}   & \rotatebox{90}{Harrier-OSS-v1-0.6B}   & \rotatebox{90}{Jina-Embeddings-v5-Text-Nano}   & \rotatebox{90}{Snowflake-Arctic-Embed-L-v2.0}   & \rotatebox{90}{Jina-Embeddings-v5-Text-Small}   & \rotatebox{90}{Pplx-Embed-v1-0.6B}   & \rotatebox{90}{\bf{EuroDense-435M}} \\ 
\hline
bBSARD-nl                                   & 16.32                                        & 24.93                             & 26.60                              & \cellcolor{silver}{32.80}       & 28.41                              & 28.79                                       & 28.83                                        & 30.90                                        & \cellcolor{bronze}{32.70}            & \cellcolor{gold}{33.08}          \\
 ArguAna-nl                                  & 47.19                                        & 53.05                             & 58.62                              & 58.41                        & \cellcolor{silver}{62.37}             & 59.15                                       & 53.30                                        & \cellcolor{bronze}{60.53}                       & 58.76                             & \cellcolor{gold}{63.09}          \\
 Climate-FEVER-nl                            & 26.78                                        & 32.48                             & 26.57                              & 20.97                        & 32.28                              & 31.36                                       & \cellcolor{silver}{34.10}                       & \cellcolor{bronze}{32.70}                       & 30.07                             & \cellcolor{gold}{35.11}          \\
\hline
\end{tabular}
\caption{\label{tab:detail_multilingaul_results_2}
Detailed evaluation results for ten multilingual models on tasks in Dutch, Russian, and English.
}
\end{table*}

\clearpage
\subsection{Evaluation of monolingual models}
\label{app:monolingual_eval}

Another experiment we conducted was a comparison of the models we developed with models designed specifically for a given language. The results presented below include up to ten models for each language, among which are the models we have proposed. Given the linguistic specifics of our models, the PolDense models were compared only in Polish, while EuroDense-435M was compared across all languages. For Spanish, Portuguese, and Italian, we did not find enough models, so the table for these languages contains fewer entries.

In Polish, we compared our models against two models from the Stella-PL \citep{dadas2024pirb} family, namely Stella-PL-Retrieval-8K and Stella-PL-Retrieval-Mini-8K, as well as the MMLW-Retrieval-RoBERTa-Large-v2 \citep{dadas2024pirb}, which achieved the highest PIRB score among MMLW models. The results for this language are presented in Table \ref{tab:pl_results}. Our PolDense-1B model proved to be the best on 26 tasks (63.41\%), outperforming existing solutions for this language. Another of our models, PolDense-400M, ranked second 23 times, outperforming the much larger Stella-PL-Retrieval-8K model on many of them. The EuroDense-435M model, designed for nine languages, achieved the best or second-best result on several tasks, outperforming four smaller PolDense models and a model from the MMLW family. The PolDense-150M model ranked third 12 times, outperforming larger models like EuroDense-435M, MMLW-Retrieval-RoBERTa-Large-v2, and both models from the Stella-PL family, in 10 of those cases.

The evaluation results for German are summarized in Table \ref{tab:eurodense_de_results}. The EuroDense-435M model outperformed the other models on four tasks; in two of them, namely LegalQuAD-de and the German part of MLDR, it significantly outperformed the second-place model, Jina-Embeddings-v2-Base-DE \citep{mohr2024}. In the other two tasks, it came in second, behind the aforementioned model from the Jina-Embeddings family and the German-English-BGE-M3\footnote{\url{https://hf.co/ferrisS/german-english-bge-m3}} bilingual model.

For the French language, the EuroDense-435M model ranked among the top 3 models for each task, winning seven of them (77.78\%), as shown in Table \ref{tab:eurodense_fr_results}. For the AlloprofRetrieval-fr, CLIRudit-fr, and French part of the MLDR, it significantly outperformed the other solutions. The French model that stood out from the rest and beat our model once is French-BGE-M3\footnote{\url{https://hf.co/antoinelouis/french-bge-m3}}. The CamemBERT-Base-lleqa  and DistilCamemBERT-lleqa models \citep{louis2023interpretable} present an interesting case; they took first and second place, respectively, on the bBSARD-fr task, noticeably outperforming the other models, including ours. This is likely because the task involves legal data, and both models were specifically trained for this domain.

Table \ref{tab:eurodense_es_results} shows a noticeable advantage of the EuroDense-435M model on nearly all tasks in Spanish. Only in one task, the Spanish subset of the MLDR dataset, our model comes in second, beaten by the Spanish-GTE-Multilingual-Base\footnote{\url{https://hf.co/CarlosRCDev/spanish-gte-multilingual-base}}. Notably, the Jina-Embeddings-v2-Base-es \citep{mohr2024} Spanish/English bilingual model ranked second in four tasks, which also earned it second position regarding the average score. The results for the other models were much worse than those for these three models.

For Italian, our model outperformed the competitors on nearly all tasks, with a particularly strong lead over other models on some, such as SQuAD-ita and the Italian parts of the MLDR and WebFAQ datasets, as shown in Table \ref{tab:eurodense_it_results}. Only on the WikipediaQA-ita task, the result was slightly worse than that of the Multilingual-E5-Large-ita\footnote{\url{https://hf.co/mik3ml/multilingual-e5-large-ita}}, which also placed second in the other tasks.

Table \ref{tab:eurodense_pt_results} shows that in Portuguese, the situation is similar to that in Italian, i.e., EuroDense-435M won all tasks except one JurisTCU-por, in which the Serafim-900m-Portuguese-pt-Sentence-Encoder-ir \citep{epia2024serafim}, which had placed second in the other tasks, proved to be the better model. Our model also demonstrated a significant advantage on several tasks here, particularly on the Portuguese MLDR subset and the Pira-por task. The other models achieved much poorer results.

Dutch is another language in which EuroDense-435M achieved significantly better results than language-specific models. Table \ref{tab:eurodense_nl_results}, which summarizes the evaluation for this language, shows that our model won 24 out of 25 tasks, losing only the LegalQA-nl task to the Dutch-English-Snowflake-Arctic-Embed-L-v2.0\footnote{\url{https://hf.co/denniscraandijk/dutch-english-snowflake-arctic-embed-l-v2.0}} model. Multiple wins across all tasks, including some by a large margin in the BEIR-nl tasks, resulted in our model outperforming the second-place E5-Large-trm-nl \citep{banar-etal-2026-mteb} model by nearly 10 percentage points on average.

In the Russian-language tasks, our model is less dominant than in the languages described earlier, although it still outperforms the others in most of them (16 out of 26), as presented in Table \ref{tab:eurodense_ru_results}. The models that surpass EuroDense-435M in Russian are primarily FRIDA\footnote{\url{https://hf.co/ai-forever/FRIDA}} and USER-BGE-M3 \citep{deepvk2024user}, which achieved the highest scores on six and three tasks, respectively.

The results in Table \ref{tab:eurodense_en_results} for English show that the EuroDense-435M model won in 12 out of 22 tasks.  If we analyze the results by the benchmarks the tasks come from, the DenseOn \citep{sourty2026denseon} model won more tasks from the BEIR benchmark, while in the RTEB benchmark, our model won definitively, winning 7 out of 8 tasks. Also deserving of mention is the Granite-Embedding-English-r2 \citep{awasthy2025graniteembeddingr2models}, which won both tasks from the scientific domain, namely SciDocs and SciFact.

\begin{table*}[!ht]
\tiny
\centering
\renewcommand{\arraystretch}{1.1}
\begin{tabular}{lrrrrrrrrrr}
\hline
 Task                                & \rotatebox{90}{\bf{PolDense-17M}}   & \rotatebox{90}{\bf{PolDense-32M}}   & \rotatebox{90}{\bf{PolDense-68M}}   & \rotatebox{90}{MMLW-Retrieval-RoBERTa-Large-v2}   & \rotatebox{90}{\bf{PolDense-150M}}   & \rotatebox{90}{\bf{EuroDense-435M}}   & \rotatebox{90}{Stella-PL-Retrieval-Mini-8K}   & \rotatebox{90}{Stella-PL-Retrieval-8K}   & \rotatebox{90}{\bf{PolDense-400M}}   & \rotatebox{90}{\bf{PolDense-1B}}   \\
\hline
 Gemini                              & 97.42                       & 98.32                       & 98.84                       & \cellcolor{silver}{99.32}                         & \cellcolor{bronze}{99.05}       & 98.91                         & \cellcolor{silver}{99.32}                     & \cellcolor{gold}{99.46}                  & 98.87                        & \cellcolor{silver}{99.32}     \\
 Odi                                 & 87.12                       & 91.23                       & 93.69                       & 94.73                                          & 94.56                        & 94.95                         & 94.98                                      & \cellcolor{bronze}{95.61}                & \cellcolor{silver}{95.82}       & \cellcolor{gold}{96.13}       \\
 Onet                                & 49.76                       & 56.05                       & 60.05                       & 62.00                                          & 62.14                        & 61.54                         & 62.45                                      & \cellcolor{bronze}{64.22}                & \cellcolor{silver}{65.64}       & \cellcolor{gold}{66.66}       \\
\hline
\end{tabular}
\caption{\label{tab:pl_results}
Detailed evaluation results for our models and language-specific models on Polish tasks.
}
\end{table*}

\begin{table*}[!ht]
\tiny
\centering
\renewcommand{\arraystretch}{1.1}
\begin{tabular}{lrrrrrrrrrr}
\hline
 Task                         & \rotatebox{90}{German\_Semantic\_V3b}   & \rotatebox{90}{GBERT-Large-Paraphrase-Euclidean}   & \rotatebox{90}{GBERT-Large-Paraphrase-Cosine}   & \rotatebox{90}{Bi-Encoder\_Msmarco\_BERT-Base\_German}   & \rotatebox{90}{German\_Semantic\_STS\_V2}   & \rotatebox{90}{German-Multilingual-E5-Small}   & \rotatebox{90}{German-English-Multilingual-E5-Small}   & \rotatebox{90}{German-English-BGE-M3}   & \rotatebox{90}{Jina-Embeddings-v2-Base-DE}   & \rotatebox{90}{\bf{EuroDense-435M}}   \\
\hline
 GerDaLIR-de                  & 6.50                                 & 6.69                                            & 6.91                                         & 8.11                                                  & \cellcolor{bronze}{9.05}                    & 7.83                                        & 7.83                                                & 7.12                                 & \cellcolor{gold}{27.98}                      & \cellcolor{silver}{26.46}        \\
 GermanDPR-de                 & 62.33                                & 66.33                                           & 64.59                                        & 75.68                                                 & 72.78                                    & 78.00                                       & 78.17                                               & \cellcolor{silver}{82.40}               & \cellcolor{bronze}{79.32}                    & \cellcolor{gold}{85.53}          \\
\hline
\end{tabular}
\caption{\label{tab:eurodense_de_results}
Detailed evaluation results for our EuroDense-435M and language-specific models on German tasks.
}
\end{table*}

\begin{table*}[!ht]
\tiny
\centering
\renewcommand{\arraystretch}{1.1}
\begin{tabular}{lrrrrrrrrrr}
\hline
 Task                 & \rotatebox{90}{DistilCamemBERT-lleqa}   & \rotatebox{90}{BiEncoder-CamemBERT-Base-MMarcoFR}   & \rotatebox{90}{CamemBERT-Base-lleqa}   & \rotatebox{90}{Bilingual-Embedding-Small}   & \rotatebox{90}{French-ME5-Base}   & \rotatebox{90}{Bilingual-Embedding-Base}   & \rotatebox{90}{French-ME5-Large}   & \rotatebox{90}{French-mGTE-Base}   & \rotatebox{90}{French-BGE-M3}   & \rotatebox{90}{\bf{EuroDense-435M}}   \\
\hline
 AlloprofRetrieval-fr & 31.28                                & 31.66                                            & 35.14                               & 39.19                                    & 36.32                          & 41.38                                   & 38.91                           & \cellcolor{silver}{48.92}          & \cellcolor{bronze}{46.78}       & \cellcolor{gold}{60.24}          \\
 bBSARD-fr            & \cellcolor{silver}{57.20}               & 13.21                                            & \cellcolor{gold}{63.10}                & 11.26                                    & 15.60                          & 15.34                                   & 18.88                           & 22.93                           & 23.60                        & \cellcolor{bronze}{34.92}        \\
 Belebele-fra\_Latn   & 81.53                                & 85.98                                            & 80.13                               & 91.16                                    & 92.14                          & \cellcolor{bronze}{94.18}                  & 93.56                           & 91.91                           & \cellcolor{silver}{94.47}       & \cellcolor{gold}{96.88}          \\
\hline
\end{tabular}
\caption{\label{tab:eurodense_fr_results}
Detailed evaluation results for our EuroDense-435M and language-specific models on French tasks.
}
\end{table*}

\begin{table*}[!ht]
\tiny
\centering
\renewcommand{\arraystretch}{1.1}
\begin{tabular}{lrrrrrrrrrr}
\hline
 Task    & \rotatebox{90}{Sentence\_Similarity\_Spanish\_es}   & \rotatebox{90}{STSb-m-mt-ES-DistilUSE-Base-Multilingual-Cased-v1}   & \rotatebox{90}{Paraphrase-Spanish-DistilRoBERTa}   & \rotatebox{90}{Spanish-GTE-Multilingual-Base}   & \rotatebox{90}{Jina-Embeddings-v2-Base-es}   & \rotatebox{90}{\bf{EuroDense-435M}}   \\
\hline
 MessIRv & 39.40                                            & 51.89                                                            & 51.68                                           & \cellcolor{bronze}{80.40}                       & \cellcolor{silver}{82.36}                    & \cellcolor{gold}{86.59}          \\
 MFAQ-es & 39.77                                            & 41.75                                                            & 48.85                                           & \cellcolor{bronze}{59.24}                       & \cellcolor{silver}{64.47}                    & \cellcolor{gold}{70.86}          \\
 MKQA-es & 4.48                                             & 4.56                                                             & 4.79                                            & \cellcolor{silver}{10.79}                       & \cellcolor{bronze}{7.92}                     & \cellcolor{gold}{12.91}          \\
 Pres    & 38.01                                            & 31.08                                                            & 33.51                                           & \cellcolor{bronze}{40.52}                       & \cellcolor{silver}{48.87}                    & \cellcolor{gold}{51.96}          \\
 SQAC    & 63.61                                            & 62.49                                                            & 65.76                                           & \cellcolor{bronze}{81.63}                       & \cellcolor{silver}{84.47}                    & \cellcolor{gold}{87.81}          \\
\hline
\end{tabular}
\caption{\label{tab:eurodense_es_results}
Detailed evaluation results for our EuroDense-435M and language-specific models on Spanish tasks.
}
\end{table*}

\begin{table*}[!ht]
\tiny
\centering
\renewcommand{\arraystretch}{1.1}
\begin{tabular}{lrrrrrrrrrr}
\hline
Task                                        & \rotatebox{90}{SPLADE-BERT-Base-Italian-XXL-Uncased-cv}   & \rotatebox{90}{Sentence-BERT-Base}   & \rotatebox{90}{Sentence-BERT-Base-Italian-XXL-Uncased}   & \rotatebox{90}{MMarco-BERT-Base-Italian-Uncased}   & \rotatebox{90}{MMarco-Sentence-BERTino}   & \rotatebox{90}{Multi-Sentence-BERTino}   & \rotatebox{90}{Sentence-BERTino}   & \rotatebox{90}{Multilingual-E5-Large-ita}   & \rotatebox{90}{\bf{EuroDense-435M}}   \\
\hline
 Belebele-ita\_Latn                          & 61.70                                                  & 71.66                             & 77.52                                                 & 67.98                                           & 81.52                                  & 82.05                                 & \cellcolor{bronze}{89.19}          & \cellcolor{silver}{91.97}                   & \cellcolor{gold}{96.97}          \\
 MKQA-it                                     & 1.57                                                   & 3.46                              & 4.86                                                  & 4.56                                            & \cellcolor{bronze}{5.78}                  & 5.69                                  & 5.66                            & \cellcolor{silver}{9.62}                    & \cellcolor{gold}{13.08}          \\
 WebFAQ-retrieval-ita                        & 25.76                                                  & 23.14                             & 37.53                                                 & 48.42                                           & 62.28                                  & \cellcolor{bronze}{63.16}                & 56.02                           & \cellcolor{silver}{65.62}                   & \cellcolor{gold}{80.88}          \\
\hline
\end{tabular}
\caption{\label{tab:eurodense_it_results}
Detailed evaluation results for our EuroDense-435M and language-specific models on Italian tasks.
}
\end{table*}

\begin{table*}[!ht]
\tiny
\centering
\renewcommand{\arraystretch}{1.1}
\begin{tabular}{lrrrrrrrrrr}
\hline
 Task                 & \rotatebox{90}{Legal-BERTimbau-Large}   & \rotatebox{90}{BERT-Large-Portuguese-Cased}   & \rotatebox{90}{BERT-Base-Portuguese-Cased-Finetuned-TCU-Acordaos}   & \rotatebox{90}{Legal-BERTimbau-sts-Large-ma-v3}   & \rotatebox{90}{Msmarco-DistilBERT-Base-tas-b-MMarco-pt-300k}   & \rotatebox{90}{BERT-Large-Portuguese-Cased-Legal-MLM-STS-v1.0}   & \rotatebox{90}{Serafim-900m-Portuguese-pt-Sentence-Encoder-ir}   & \rotatebox{90}{\bf{EuroDense-435M}}   \\
\hline
 Belebele-por\_Latn   & 47.33                                & 59.88                                      & 63.05                                                            & 69.49                                          & 52.59                                                       & \cellcolor{bronze}{75.56}                                        & \cellcolor{silver}{89.64}                                        & \cellcolor{gold}{97.03}          \\
 JurisTCU-por         & 22.53                                & 19.44                                      & 18.06                                                            & 20.64                                          & 25.68                                                       & \cellcolor{bronze}{35.66}                                        & \cellcolor{gold}{56.42}                                          & \cellcolor{silver}{53.02}        \\
 MFAQ-pt              & 28.07                                & 33.71                                      & 29.35                                                            & 33.02                                          & 31.45                                                       & \cellcolor{bronze}{46.76}                                        & \cellcolor{silver}{63.71}                                        & \cellcolor{gold}{72.88}          \\
\hline
\end{tabular}
\caption{\label{tab:eurodense_pt_results}
Detailed evaluation results for our EuroDense-435M and language-specific models on Portugues tasks.
}
\end{table*}

\begin{table*}[!ht]
\tiny
\centering
\renewcommand{\arraystretch}{1.1}
\begin{tabular}{lrrrrrrrrrr}
\hline
 Task                                        & \rotatebox{90}{RobBERT-2023-Large-ft}   & \rotatebox{90}{E5-Large-v2-t2t-nl}   & \rotatebox{90}{E5-Large-v2-t2t}   & \rotatebox{90}{RobBERT-2023-Base-ft}   & \rotatebox{90}{E5-Small-trm-nl}   & \rotatebox{90}{E5-Base-trm-nl}   & \rotatebox{90}{Dutch-English-Snowflake-Arctic-Embed-L-v2.0}   & \rotatebox{90}{Dutch-Multilingual-E5-Large}   & \rotatebox{90}{E5-Large-trm-nl}   & \rotatebox{90}{\bf{EuroDense-435M}}   \\
\hline
 bBSARD-nl                                   & 18.99                                & 16.35                             & 16.91                          & 17.60                               & 11.07                          & 17.81                         & \cellcolor{silver}{25.16}                                     & 21.05                                      & \cellcolor{bronze}{22.40}         & \cellcolor{gold}{33.08}          \\
 ArguAna-nl                                  & 46.15                                & 43.48                             & 39.52                          & 44.05                               & 48.32                          & 49.71                         & \cellcolor{silver}{53.77}                                     & \cellcolor{bronze}{50.38}                     & 49.79                          & \cellcolor{gold}{63.09}          \\
 Climate-FEVER-nl                            & 21.79                                & 17.64                             & 17.68                          & \cellcolor{bronze}{25.02}              & 18.04                          & 15.99                         & \cellcolor{silver}{26.60}                                     & 15.33                                      & 18.42                          & \cellcolor{gold}{35.11}          \\
\hline
\end{tabular}
\caption{\label{tab:eurodense_nl_results}
Detailed evaluation results for our EuroDense-435M and language-specific models on Dutch tasks.
}
\end{table*}

\begin{table*}[!ht]
\tiny
\centering
\renewcommand{\arraystretch}{1.1}
\begin{tabular}{lrrrrrrrrrr}
\hline
 Task                 & \rotatebox{90}{BERTA}     & \rotatebox{90}{Giga-Embeddings-Instruct}   & \rotatebox{90}{USER2-Small}   & \rotatebox{90}{LaBSE-ru-turbo}   & \rotatebox{90}{RuBERT-mini-frida}   & \rotatebox{90}{ru-EN-RoSBERTa}   & \rotatebox{90}{USER2-Base}   & \rotatebox{90}{USER-BGE-M3}   & \rotatebox{90}{FRIDA}     & \rotatebox{90}{\bf{EuroDense-435M}}   \\
\hline
 Belebele-rus\_Cyrl   & 81.87                  & 85.69                                   & 89.31                      & \cellcolor{bronze}{92.55}        & 90.46                            & 91.93                         & 92.50                     & \cellcolor{silver}{93.23}     & 89.91                  & \cellcolor{gold}{97.24}          \\
 MFAQ-ru              & 63.19                  & 74.09                                   & 69.15                      & 70.57                         & 70.10                            & 73.50                         & 74.78                     & \cellcolor{bronze}{76.01}     & \cellcolor{silver}{76.15} & \cellcolor{gold}{78.87}          \\
 MKQA-ru              & 4.86                   & \cellcolor{silver}{10.20}                  & 5.91                       & 4.56                          & 4.63                             & \cellcolor{bronze}{7.10}         & 6.92                      & 6.32                       & 6.45                   & \cellcolor{gold}{12.65}          \\
 Ria-News             & 77.54                  & \cellcolor{bronze}{80.43}                  & 74.30                      & 69.33                         & 72.05                            & 79.21                         & 79.37                     & \cellcolor{gold}{83.47}       & \cellcolor{silver}{82.02} & 78.86                         \\
\hline
\end{tabular}
\caption{\label{tab:eurodense_ru_results}
Detailed evaluation results for our EuroDense-435M and language-specific models on Russian tasks.
}
\end{table*}

\begin{table*}[!ht]
\tiny
\centering
\renewcommand{\arraystretch}{1.1}
\begin{tabular}{lrrrrrrrrrr}
\hline
 Task                  & \rotatebox{90}{E5-Small-v2}   & \rotatebox{90}{Granite-Embedding-Small-English-r2}   & \rotatebox{90}{E5-Base-v2}   & \rotatebox{90}{BGE-Small-EN-v1.5}   & \rotatebox{90}{E5-Large-v2}   & \rotatebox{90}{Granite-Embedding-English-r2}   & \rotatebox{90}{BGE-Base-EN-v1.5}   & \rotatebox{90}{BGE-Large-EN-v1.5}   & \rotatebox{90}{DenseOn}   & \rotatebox{90}{\bf{EuroDense-435M}}   \\
\hline
 ArguAna               & 41.85                      & 54.44                                             & 44.61                     & 60.45                            & 46.54                      & 59.24                                       & \cellcolor{bronze}{63.91}          & \cellcolor{silver}{64.68}           & 54.70                  & \cellcolor{gold}{65.08}          \\
 Climate-FEVER         & 22.85                      & 31.73                                             & 26.60                     & 31.83                            & 22.18                      & \cellcolor{bronze}{35.84}                      & 31.18                           & \cellcolor{silver}{36.56}           & \cellcolor{gold}{37.58}   & 35.75                         \\
 DBPedia-entity        & 41.31                      & 38.04                                             & 42.20                     & 40.03                            & 44.01                      & 39.75                                       & 40.77                           & \cellcolor{bronze}{44.11}           & \cellcolor{silver}{44.57} & \cellcolor{gold}{46.35}          \\
 FEVER                 & 81.63                      & 86.51                                             & 84.98                     & 86.63                            & 82.81                      & \cellcolor{silver}{88.09}                      & 86.30                           & \cellcolor{bronze}{87.19}           & \cellcolor{gold}{90.80}   & 85.30                         \\
\hline
\end{tabular}
\caption{\label{tab:eurodense_en_results}
Detailed evaluation results for our EuroDense-435M and language-specific models on English tasks.
}
\end{table*}

\begin{table*}[!ht]
\tiny
\centering
\renewcommand{\arraystretch}{1.1}
\begin{tabular}{lll}
\hline
\bf{Model} & \bf{Hugging Face model name} & \bf{Reference} \\
\hline
Qwen3-Embedding-4B & Qwen/Qwen3-Embedding-4B & \citep{zhang2025qwen3} \\
Qwen3-Embedding-8B & Qwen/Qwen3-Embedding-8B & \citep{zhang2025qwen3} \\
Pplx-Embed-v1-4b & perplexity-ai/pplx-embed-v1-4b &  \citep{eslami2026diffusion}\\
BGE-Multilingual-Gemma2 & BAAI/bge-multilingual-gemma2 & \citep{chen-etal-2024-m3, xiao2024bge} \\
Inf-Retriever-v1 & infly/inf-retriever-v1 & \citep{infly-ai_2025} \\
\hline
Multilingual-E5-Small & intfloat/multilingual-e5-small & \citep{wang2024multilingual} \\
Multilingual-E5-Base & intfloat/multilingual-e5-base & \citep{wang2024multilingual} \\
\hline
\end{tabular}
\caption{\label{tab:models}
A list of external models evaluated in this publication for comparison with ours.
}
\end{table*}

\end{document}